\documentclass[conference]{IEEEtran}
\IEEEoverridecommandlockouts
\usepackage{cite}
\usepackage{amsmath,amssymb,amsfonts}
\usepackage{algorithmic}
\usepackage{graphicx}
\usepackage{textcomp}
\usepackage{xcolor}

\usepackage[center]{subfigure}
\usepackage{multirow}
\usepackage{mathtools}
\graphicspath{{Figures/}}

\def\BibTeX{{\rm B\kern-.05em{\sc i\kern-.025em b}\kern-.08em
    T\kern-.1667em\lower.7ex\hbox{E}\kern-.125emX}}

\begin{document}

\title{\vspace{18pt}Tactile Hallucinations on Artificial Skin Induced by Homeostasis in a Deep Boltzmann Machine}

\author{Michael Deistler$^{*}$,  Yagmur Yener$^{*}$, Florian Bergner, Pablo Lanillos, and Gordon Cheng
\thanks{$^{*}$ Authors contributed equally}
\thanks{All authors are with the Institute for Cognitive Systems (ICS),  Technische Universit\"at M\"unchen, Arcisstraße 21 80333 M\"unchen, Germany
        {\tt\small (www.ics.ei.tum.de)}}%
\thanks{M. Deistler and  Yagmur Yener were supported by the Elite Master of Science program in Neuroengineering (MSNE). P. Lanillos was supported by SELFCEPTION project (www.selfception.eu) European Union Horizon 2020 Programme under grant agreement n. 741941}}

\maketitle

\begin{abstract}
Perceptual hallucinations are present in neurological and psychiatric disorders and amputees. While the hallucinations can be drug-induced, it has been described that they can even be provoked in healthy subjects. Understanding their manifestation could thus unveil how the brain processes sensory information and might evidence the generative nature of perception. In this work, we investigate the generation of tactile hallucinations on biologically inspired, artificial skin. To model tactile hallucinations, we apply homeostasis, a change in the excitability of neurons during sensory deprivation, in a Deep Boltzmann Machine (DBM). We find that homeostasis prompts hallucinations of previously learned patterns on the artificial skin in the absence of sensory input. Moreover, we show that homeostasis is capable of inducing the formation of meaningful latent representations in a DBM and that it significantly increases the quality of the reconstruction of these latent states. Through this, our work provides a possible explanation for the nature of tactile hallucinations and highlights homeostatic processes as a potential underlying mechanism.
\end{abstract}


\section{Introduction}
\label{sec:intro}

Hallucinations have been observed in several sensory systems, yet the underlying mechanisms are still poorly understood. In the somatosensory system, different kinds of hallucinations can emerge. In patients with an amputated arm or leg, phantom limb sensations have been described in the position of the missing limb \cite{rabins1994genesis, pang2016hallucinations}. Tactile hallucinations have also been described in people with neuropsychiatric diseases, most notably schizophrenia or Parkinson's disease \cite{fenelon2002tactile, mueser1990hallucinations}. In order to treat these diseases, a clear understanding of the underlying mechanisms is required. The clinical heterogeneity of hallucinations and their potential variety of underlying causes display a great challenge and opportunity from the computational modelling point of view \cite{lanillos2019NN_SZ_ASD}. Developing mathematical models that are able to explain these effects provides answers to how the brain perceives the world. In fact, it has been recently shown that hallucinations can be even provoked in healthy subjects \cite{powers2017pavlovian} by means of sensory contingency conditioning. 

Previous modelling works have focused on auditory \cite{hoffman2001book} and visual hallucinations \cite{series2010hallucinations}. However, fewer models like \cite{spitzer1995neural,brown2013active} have properly addressed tactile hallucinations despite its implications on prosthetic \cite{bostrom2014computational} and embodiment disorders.

Here, we provide such a framework inspired by recent work on the visual system. Therein, the emergence of visual hallucinations in visually impaired individuals, known as the Charles-Bonnet-Syndrome (CBS) \cite{menon2003complex}, was modelled by applying homeostasis in a Deep Boltzmann Machine (DBM) \cite{series2010hallucinations}. Homeostasis is the adjustment of the excitability of the neurons due to reduced activity. This model provides a possible explanation for several observations made in people with CBS. As the CBS is frequently compared to the phantom limb or other kinds of tactile hallucinations \cite{schultz1991charles, menon2003complex}, we here extend the existing approach to tactile hallucinations using a previously developed artificial skin \cite{Mittendorfer2011}. Our neuro-inspired framework provides a possible explanation for the emergence of tactile hallucinations and gives insights into the neurological processes during neuropsychiatric diseases such as Parkinson's or schizophrenia.

We first explain our approach which uses a DBM to model tactile hallucinations (section \ref{sec:model}). Secondly, the experimental setup and the usage of the artificial skin for creating tactile patterns and visualizing hallucinations is presented in section \ref{sec:skin}. Thirdly, we show our main results where homeostasis in a DBM leads to tactile hallucinations on an artificial skin (section \ref{sec:results}). We conclude by a discussion of the neuroscientific relevance of this work and its potential impact in medical and engineering applications (section \ref{sec:discussion}).

\section{Proposed model}
\label{sec:model}

\subsection{Modelling tactile hallucinations}
\label{sec:intro:models}
Few works in the literature have proposed computational models for hallucinations \cite{lanillos2019NN_SZ_ASD,jardri2013computational}. Auditory hallucinations of schizophrenic patients \cite{hoffman2001book,ruppin1995neural} and visual hallucinations for the Charles-Bonnet syndrome \cite{series2010hallucinations} were addressed by means of neural networks. Furthermore, perceptual hallucinations in a Bayesian framework were discussed in \cite{adams2013computational} under the free-energy principle and in \cite{jardri2013computational} based on the circular inference hypothesis.

\begin{figure}
    \centering
    \includegraphics[width=.9\linewidth]{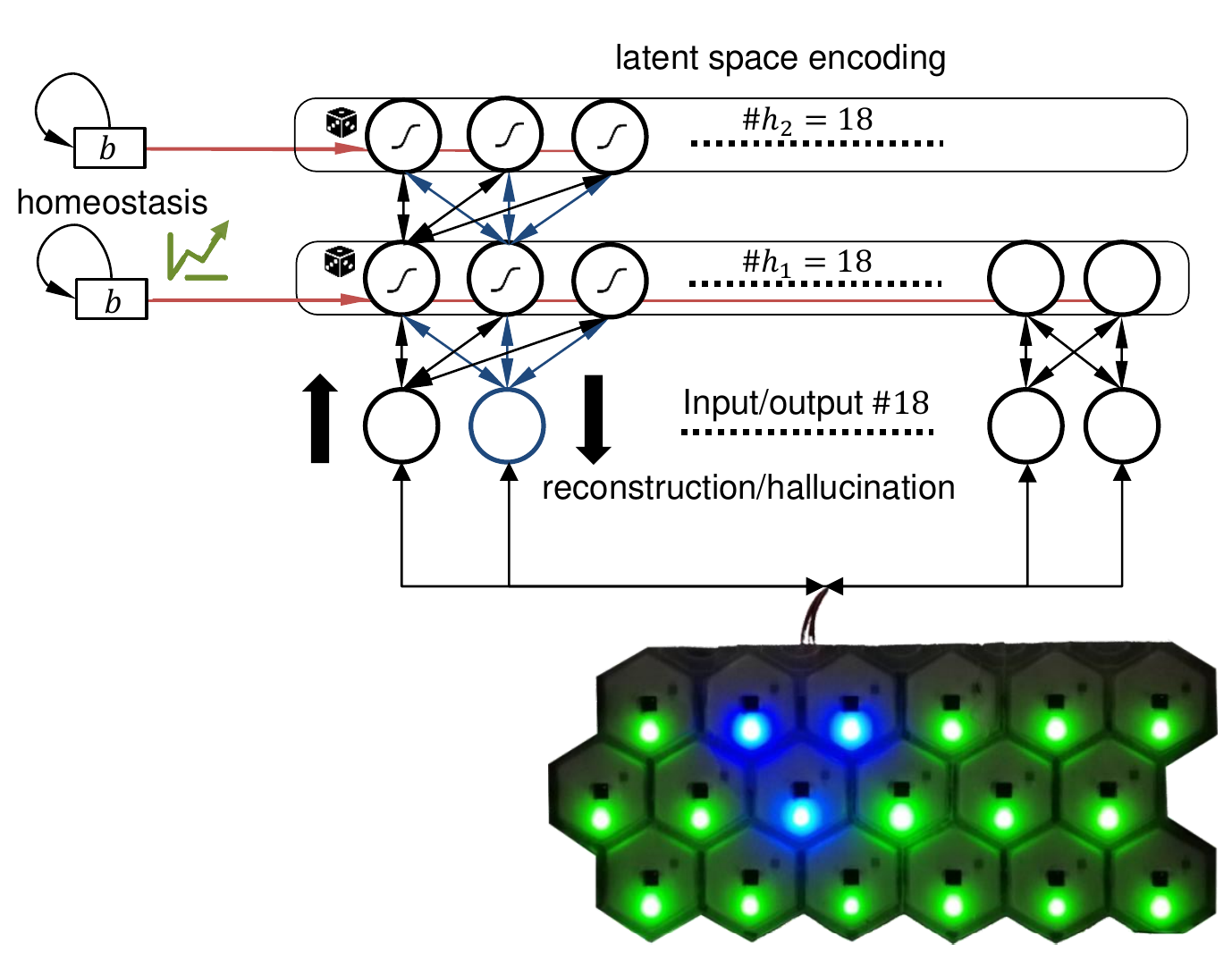}
    \caption{Deep Boltzmann Machine (DBM) for modelling tactile hallucinations. Each of the 18 artificial skin cells is connected to an input neuron. Input neurons are connected by receptive fields to the hidden layers with sigmoidal activation functions and binary neurons. In the deepest hidden layer $h_2$ also provides inputs to the first hidden layer $h_1$. Homeostasis is modeled as a change on the biases.}
    \label{fig:abstract}
\end{figure}

In order to model tactile hallucinations, this work leverages a Deep Boltzmann Machine (DBM), pre-trained with a Deep Belief Network (DBN) (see appendix). The overall setup of our model is depicted in Fig. \ref{fig:abstract}. The artificial skin cells provide the input to the visual layer of the DBM, which is then trained on those patterns. We will show that the DBM learns hidden representations of the input data and that these representations are encoded in the hidden layers of the network. We claim that, through homeostasis, the network will even in absence of sensory input produce meaningful latent representations corresponding to previously learned patterns, i.e. a hallucination pattern. As shown in Fig. \ref{fig:abstract}, homeostasis is modeled as an increase (or decrease) of the bias values of the network. The colored LEDs of the skin were used to visualize both the input and the output of the DBM.

\subsection{Receptive Fields}
\label{sec:recField}
Cortical connectivity has been observed to be hierarchical and sparse \cite{iwamura1998hierarchical, felleman1991distributed}. To model this, we use a limited connectivity between the layers of the DBM, where each neuron is connected only to neurons in the same or neighbouring columns (comparable to cortical columns), as depicted in Fig. \ref{fig:abstract}. In order to allow for the same number of input connections for all neurons in a certain layer, we also consider a ring-like connectivity. Here, every neuron is again connected only to neurons in the same or neighbouring columns. Unlike in the first case, the columns are now ordered in a ring-like, circular, structure. We refer to the two described cases as \textit{linear} and \textit{circular}.

\subsection{Homeostasis}
\label{sec:homeostasis}
At the heart of our approach lies the change of excitability of neurons during sensory deprivation. On a single neuron scale, this process is termed homeostasis. Its timescale is on the order of hours or days \cite{turrigiano2008self, turrigiano2017dialectic, watt2010homeostatic}. In general, homeostatic mechanisms decrease the excitability of highly active neurons and increase the excitability of inactive neurons. It is thus often considered to be underlying stable network function and to prevent runaway network excitability in Hebbian cell assemblies \cite{zenke2013synaptic}. Here, we make use of homeostatic mechanisms to evoke hallucinations corresponding to learned patterns. We do this by measuring the average activity of each neuron when presented with patterns from the training dataset. This average $\mu$ is then considered to be the healthy activity of that neuron. When presenting a zeros vector as an input of the network, the activity of the neuron $a_i$ will deviate from this baseline. In our model, homeostatic mechanisms will increase (or decrease) the bias $b_i$ of neuron $i$ in order to regain a healthy activity level:
\begin{equation}
    \Delta b_i = \eta (\mu_i - a_i),
\end{equation}
where $\eta$ is the adaptation rate, set to 0.01 for all neurons. Note that, due to the symmetry of the DBM, homeostatic mechanisms can both increase or decrease the bias in response to blank input.

\subsection{Decoding the hidden state}
\label{sec:decoding}
In order to evaluate the hidden states of the DBM, we need a method to decode the internal states of the hidden layers and to infer whether the hidden representations are mere noise or whether they correspond to a pattern. Instead of training a classifier on the states of the hidden units, we use the DBM as its own decoder \cite{series2010hallucinations}.

In this process, we pass the hidden state through the network towards the visible layer in a single feedforward pass. While connections in a DBM are usually bidirectional, in this feedfordward pass, all neurons in a layer get input only from the adjacent deeper layer. Therefore, the total input to a neuron decreases due to the lack of input from the adjacent shallower layer. We compensate for this lack of input by multiplying the weights by a factor of two. The weights attached to the visible layer, however, are not multiplied by two as the visible units always get input only from one side. It is important to note that we apply this process only to decode the internal states of the network for inspection. This process is hence not required to be biologically plausible. To draw a comparison to biological neural networks, this process would correspond to matching measured neuronal activity (e.g. from local field potential) to the prevalent external stimulus.

\subsection{Evaluation measure}
\label{sec:eval}
In order to evaluate that the hidden representations correspond to patterns that come from the same distribution as the dataset, we analyzed the quality of the reconstructions by means of the Dice-coefficient $D$
\begin{equation}
    D(A,B) = \frac{2|A \cap B|}{|A| + |B|},
\end{equation}
where $A$ and $B$ are two binary patterns and $|\cdot|$ denotes the cardinality of the patterns (or, in other words, the number of values that are one).

Thus, given the patterns in the dataset $P_i$ and the pattern to be evaluated $S$, we define the performance of the network $Q$ as the maximal Dice-coefficient between $S$ and any of the patterns in the dataset
\begin{equation}
    Q = \max_{i} D(P_i, S).
\end{equation}
Intuitively, if the performance of the network is one, its output (i.e. the decoding sample or the hallucination) corresponds to one of the patterns. If the performance is zero, the output does not have an overlap with any of the patterns in the training dataset.

\section{Artificial Skin and Experimental Setup}
\label{sec:skin}

\begin{figure}
    \centering
    \includegraphics[width=.8\linewidth]{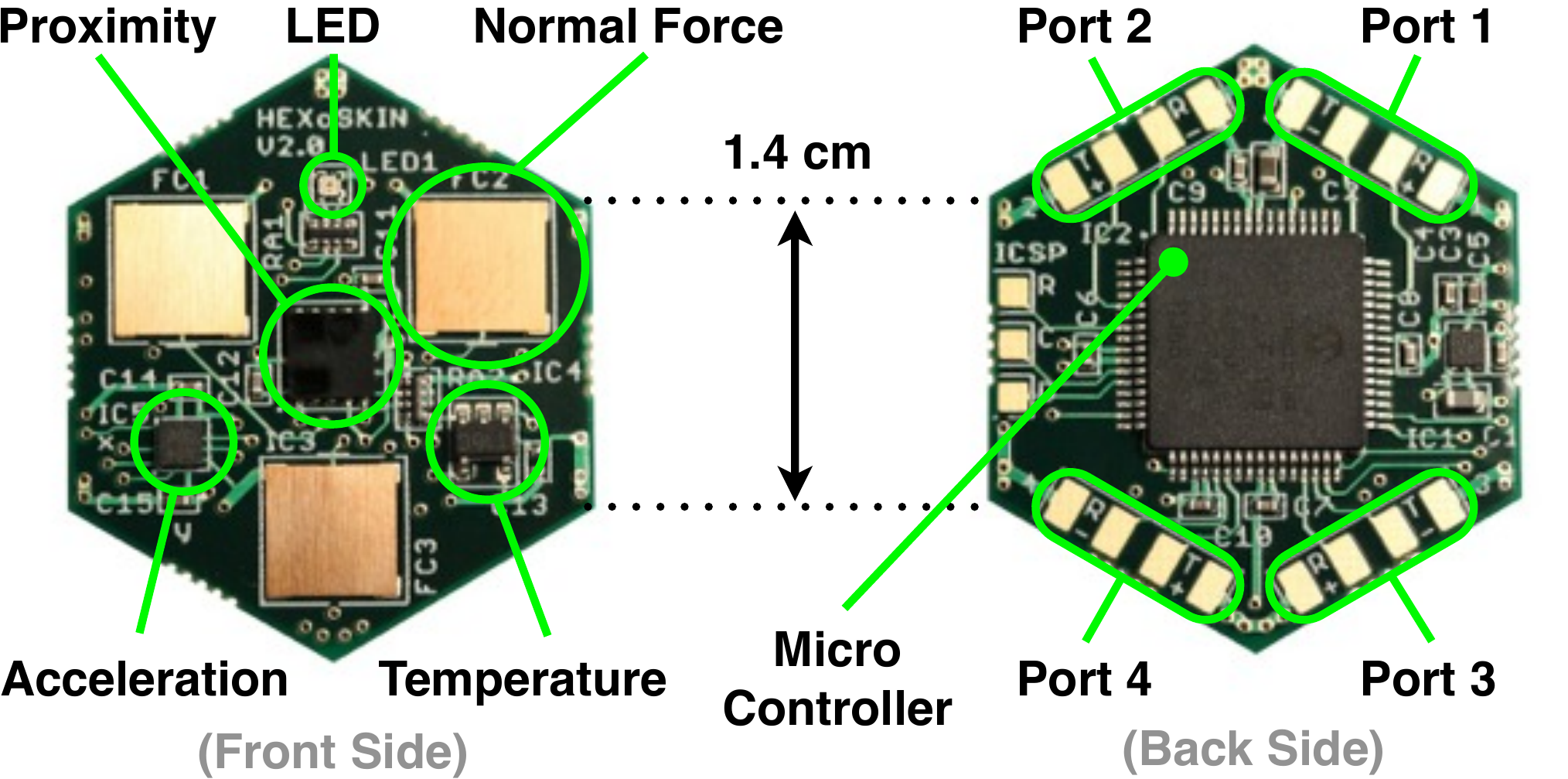}
    \caption{Biologically inspired multi-modal skin \cite{Mittendorfer2011,Bergner2016}. The skin is composed of hexagonally shaped skin cells. Each skin cell employs four different kind of sensors: three capacitive force sensors to sense contacts, one proximity sensor to sense pre-contacts, a 3D acceleration sensor to sense vibrations, and a temperature sensor. These skin cells communicate with each other, create a self-organizing communication network, and form skin patches.   
    }
    \label{fig:RobotSkin}
\end{figure}

We work with an artificial skin \cite{Mittendorfer2011,Bergner2016} consisting of 18 skin cells, which are ordered in a hexagonal grid with shape 3 $\times$ 6 (Fig \ref{fig:abstract}). Each of these cells is equipped with, among others, three normal force sensors (Fig. \ref{fig:RobotSkin}).

For the pattern acquisition, we take the maximal recorded value of the three force sensors per cell. For further robustness, we cluster $C=5$ time steps into one round. If a certain cell exceeds its force threshold $\theta_F$ within the round, it is considered to be \textit{on} in this interval. Each time step is 250 ms long, which means that one round corresponds to 1.25 seconds.

When recording, we accept the pattern for the dataset only if at least $N=2$ cells are \textit{on} in this round. This prevents having a vast amount of either empty or noisy data in the dataset. 

When the measured value does not exceed $\theta_F$, the LED lights up in green. If the respective threshold is exceeded, the LED of the respective artificial skin cell lights up in blue.

The skin connection is bidirectional. Thus, it also displays patterns extracted from the output of the DBM and shows them for $D=3$ time steps, with each time step again corresponding to 250 ms.

To model the tactile hallucinations presented in the result section \ref{sec:results}, a proof-of-concept dataset was created, consisting of three patterns with one triangle each, where a triangle is made up of three cells. The size of the patterns was selected for visual comprehension.

\begin{figure}[h]
    \centering
    \subfigure[Input procedure]{
		\includegraphics[width=0.25\columnwidth]{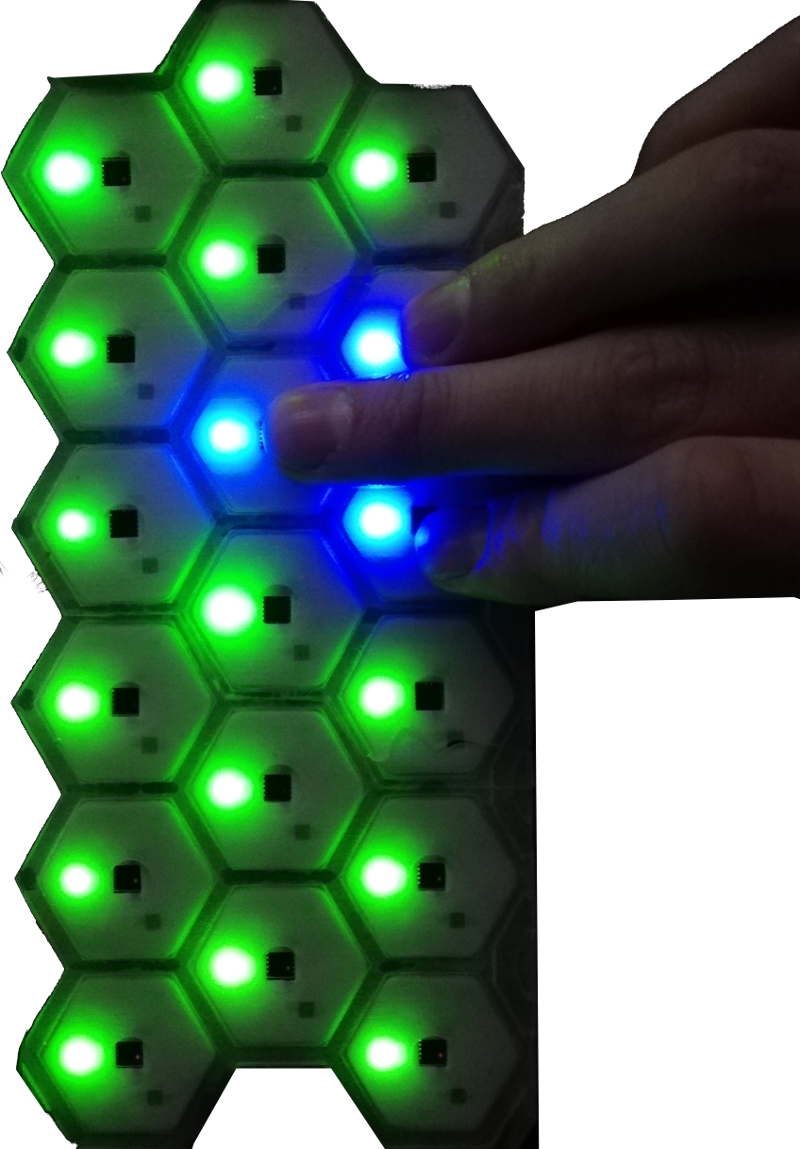}
		\label{results:exp1:error-cycle}}
	\subfigure[Examples of training patterns]{
	\includegraphics[width=0.45\columnwidth]{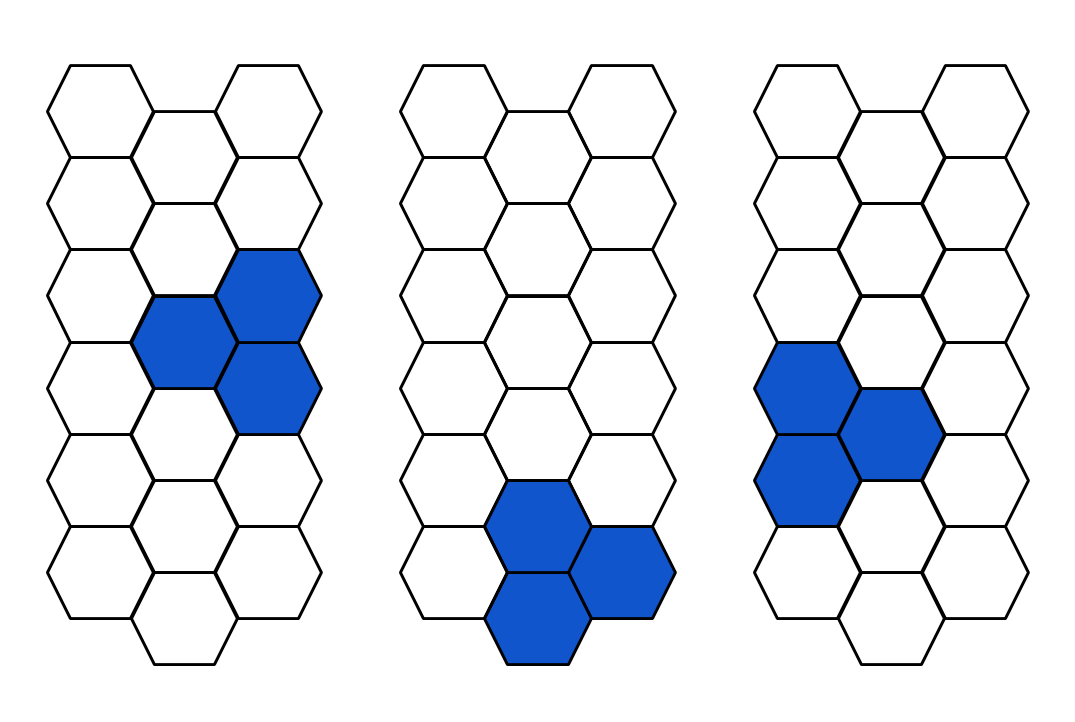}}
    \caption{Tactile patterns for training.}
    \label{fig:train_data}
\end{figure}

For reproducibility, the hyperparameters of the proposed method, and their meaning, as well as their options and a recommended default value, are given in Table \ref{tab:skin}.

\begin{table}
\caption{Overview of the hyperparameters for the artificial skin}
\label{tab:skin}       
\begin{tabular}{llll}
\hline\noalign{\smallskip}
Hyperparameter & Variable & Options & Value\\
\noalign{\smallskip}\hline\noalign{\smallskip}
MAX\_FORCE             & $\theta_F$  & [0,1]                   & 0.012 \\
MIN\_NUMBER\_OF\_CELLS & $N$         & $\in \mathbb{N}$        & 2     \\
COMBINE\_ITER          & $C$         & $\in \mathbb{N}$        & 5     \\
DISPLAY\_DURATION      & $D$         & $\in \mathbb{N}$        & 3     \\
\noalign{\smallskip}\hline
\end{tabular}
\end{table}

\section{Results}
\label{sec:results}

\subsection{Training and sampling from the DBM}
\label{sec:trainDBMresults}
\begin{figure}[h]
    \centering
    \includegraphics[width=0.48\textwidth]{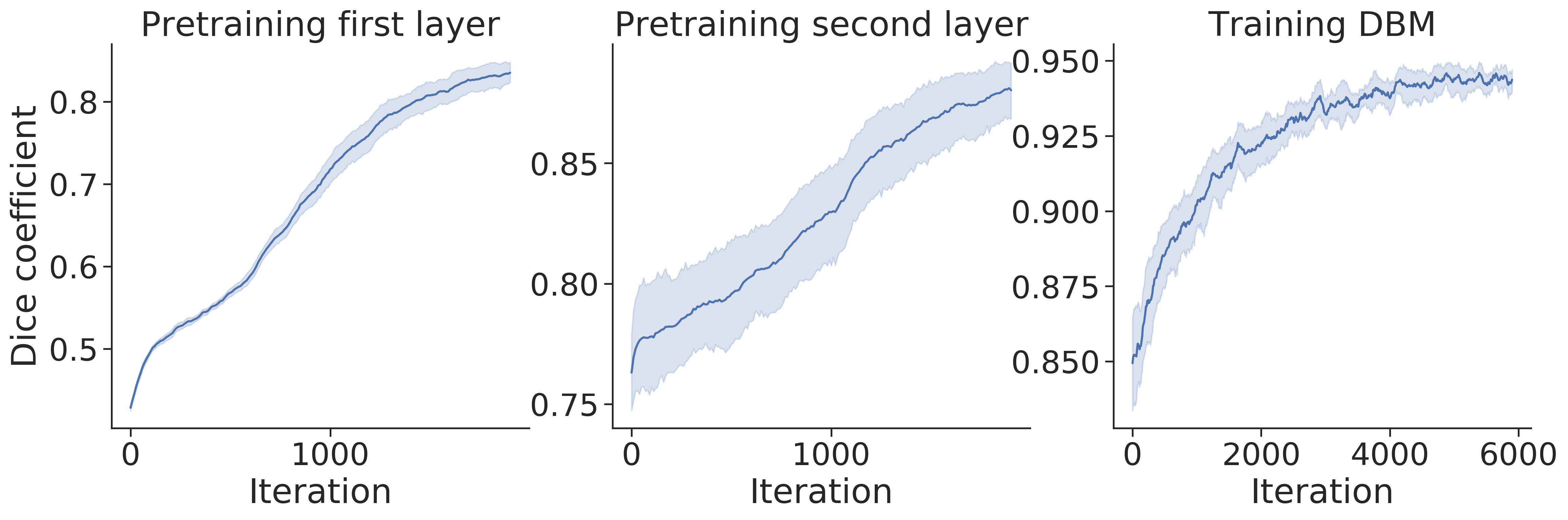}
    \caption{Training accuracy curves as measured by the performance measure $Q$ (section \ref{sec:eval}), averaged over ten trials (solid line represents mean and shaded area represents standard deviation). First, the two layers are trained separately (left and middle). Then, the pre-trained weights are transferred to a DBM. After that, the DBM is trained (right).}
    \label{fig:curves}
\end{figure}

For the below described results, we used a DBM with three layers, with each layer consisting of 18 neurons. During training, samples from the network were evaluated by calculating the Dice-coefficient-based evaluation measure $Q$ (see section \ref{sec:eval}). We pretrained the DBM as a Deep Belief Network. Thus, the training process of a three-layer DBM has three phases. In the first phase, the weights between the first and the second layer are pre-trained. After 2000 iterations, a performance $Q$ of around 0.85 is reached (Fig. \ref{fig:curves}, left). In the second phase, only the weights between the second and the third layer are pre-trained by taking the latent representations in the first hidden layer as training data. Again, the performance $Q$ reaches a value of around 0.85. Lastly, samples from the entire DBM were taken. The pre-training accuracy of the DBM is already around 0.85. The last phase is to train the overall DBM, refining the network to a performance $Q$ of 0.97. We apply early stopping to ensure that the network does not collapse onto only one of the patterns.

\subsection{Decoding the latent representations}
\begin{figure}[h]
    \centering
    \includegraphics[width=0.45\textwidth]{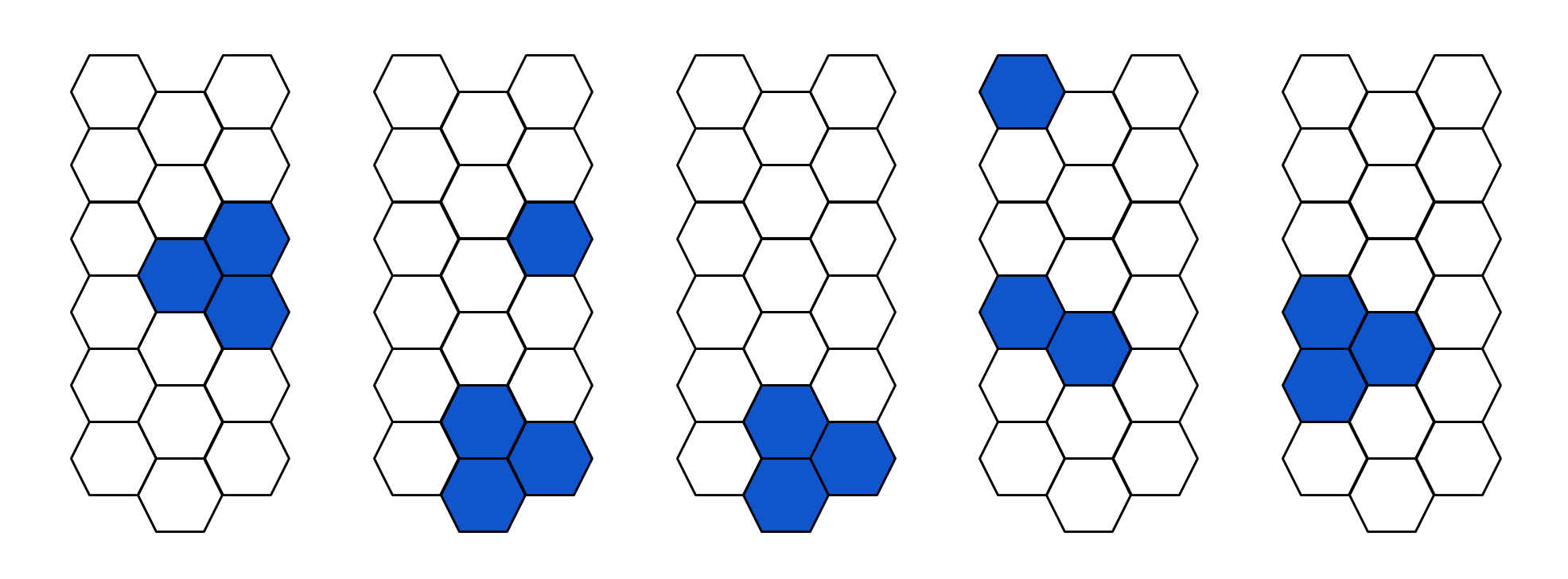}
    \caption{Decoded states when clamping to one of the training patterns for \textit{circular} connectivity. The measure of performance $Q$ is 0.87.}
    \label{fig:samples}
\end{figure}

Once the DBM has been trained, it is able to create samples from the training data. To inspect whether the DBM also forms meaningful latent representations, we decode the hidden states of the trained DBM using the decoding procedure described in section \ref{sec:decoding}. The decoding method was tested in three scenarios. As in the training, the quality of the decoding was evaluated using the measure $Q$. In the first scenario, the DBM is clamped to one of the patterns. In this case, the hidden representations in the second hidden layer clearly represent the data the network is clamped to. Thus, the decoded states mostly represent one of the training patterns, as shown in figure \ref{fig:samples}. The performance $Q_{pattern}$ for \textit{circular} connectivity is about 0.87 in this case.

As a second scenario, the input pattern was corrupted by turning two of the three cells off. The DBM is then, again, clamped to this input and the deepest layer is decoded. By corrupting the pattern, the performance $Q_{corrupted}$ of the decoded pattern dropped to about 0.64 for \textit{cicurlar} connectivity.

Lastly, fully empty patterns were fed to the network. As a result, the performance drops even further and is now only around $Q_{blank}=0.50$ for \textit{cicurlar} connectivity. 

\begin{figure}[h]
    \centering
    \includegraphics[width=0.45\textwidth]{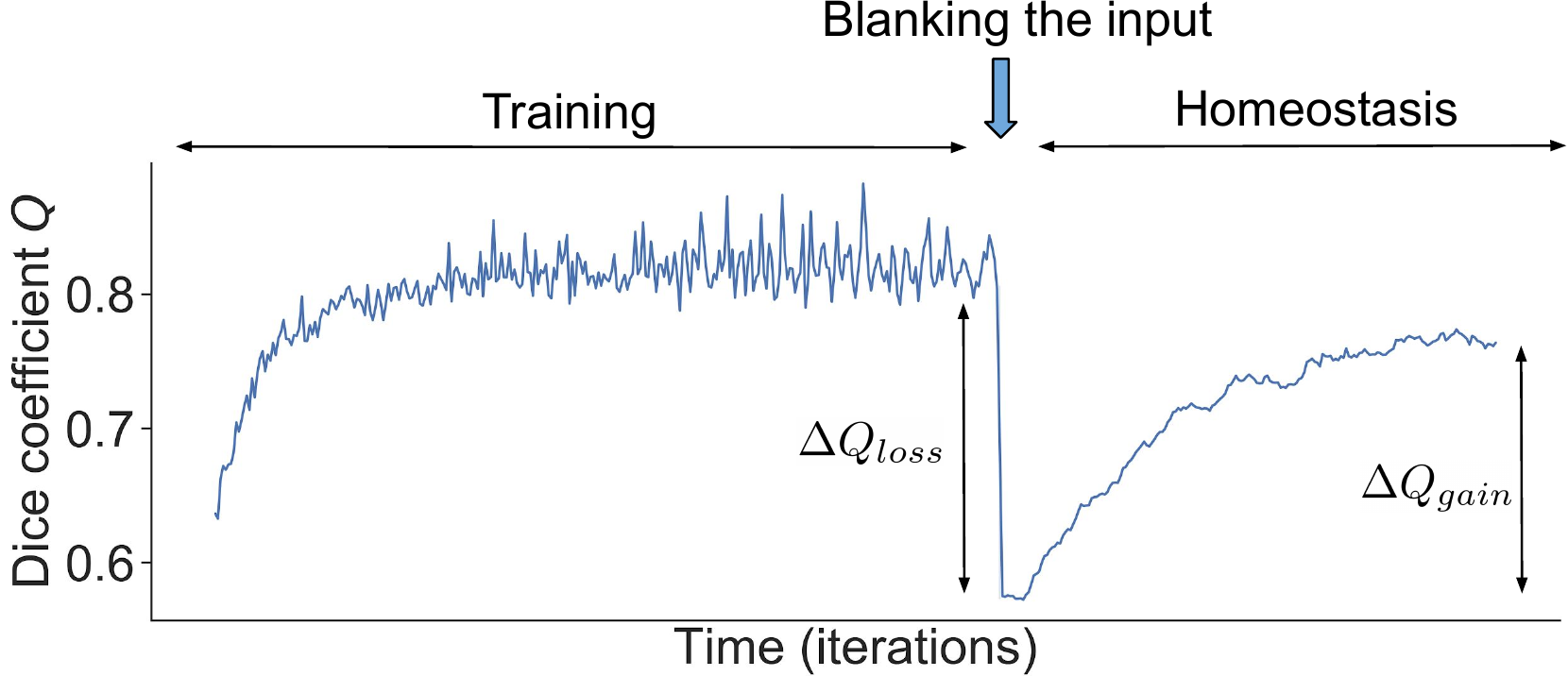}
    \caption{Visualization of the measures $\Delta Q_{loss}$ and $\Delta Q_{gain}$. In the training phase of the DBM (corresponding to figure \ref{fig:curves}, right), the accuracy of the decoding goes up to around 0.87. Then, we present the network with blank tactile patterns. The drop in decoding accuracy is termed $\Delta Q_{loss}$. After that, homeostatic processes start. The gain in performance through homeostasis is termed $\Delta Q_{gain}$.}
    \label{fig:q_loss_q_gain}
\end{figure}

Table \ref{tab:results} shows the performance of the model for the different scenarios. For further analysis, we also define the measure $\Delta Q_{loss}$, which is the difference in performances between the presentation of training patterns and zeros input - see Fig. \ref{fig:q_loss_q_gain}.
\begin{equation}
    \Delta Q_{loss} = Q_{pattern} - Q_{blank}
\end{equation}

\begin{table}[h]
\centering
\caption{Accuracy of the decoded reconstructions of the hidden layer representations. All values are the mean performance over ten trials.}
\label{tab:results}       
\begin{tabular}{llll}
\hline\noalign{\smallskip}
Receptive field & \textit{circular} & \textit{linear}\\
\noalign{\smallskip}\hline\noalign{\smallskip}
$Q_{pattern}$       & 0.87 & 0.83 \\
$Q_{corrupted}$     & 0.58 & 0.50 \\
$Q_{blank}$         & 0.50 & 0.42 \\
\noalign{\smallskip}\hline\noalign{\smallskip}
$\Delta Q_{loss}$ & 0.37  & 0.41\\
\noalign{\smallskip}\hline
\hline\noalign{\smallskip}
$Q_{hallucination}$  & 0.72  & 0.62\\
\noalign{\smallskip}\hline\noalign{\smallskip}
$\Delta Q_{gain}$ & 0.22  & 0.20\\
\noalign{\smallskip}\hline
\end{tabular}
\end{table}

\subsection{Homeostasis causes hallucinations}

We have shown that presenting the network with a corrupted or blank input impedes the reconstruction quality. Now we show that, when introducing the homeostatic processes described in section \ref{sec:homeostasis}, the performance $Q$ of the network is increased when presented with a blank pattern and meaningful latent states are retrieved.

\begin{figure}[h]
\centering
\includegraphics[width=58mm]{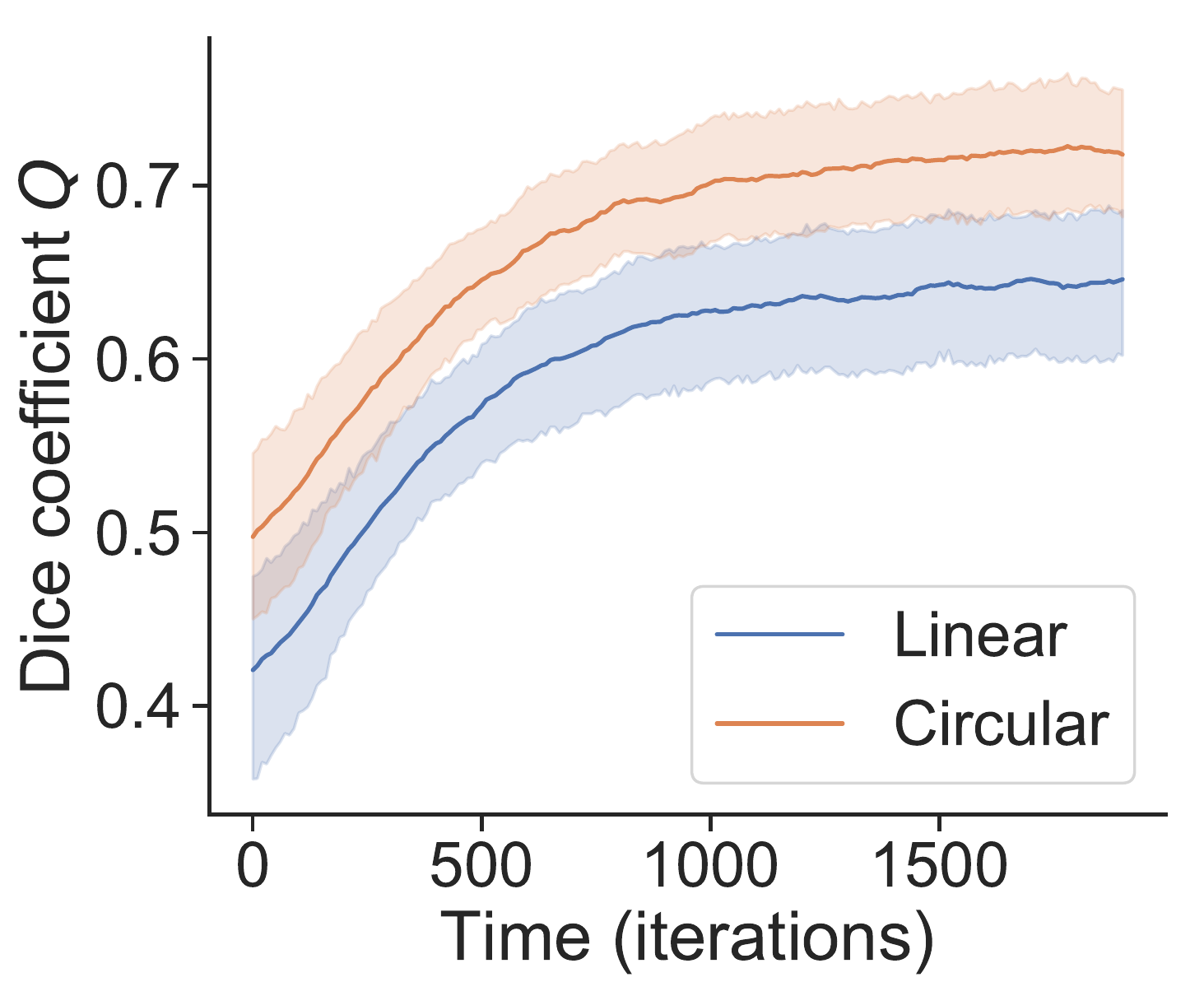}
\caption{Increase in reconstruction performance through homeostasis. Reconstruction quality $Q$ over time, averaged over ten trials, corresponding to ten independently trained DBMs. After each time step, the biases are updated and get closer to the baseline level. The two traces correspond to the two types of connectivity namely $\textit{circular}$ (orange), and $\textit{linear}$ (blue). In both cases, the homeostatic processes clearly increase the quality of the reconstructions. The shaded area represents the standard deviation.}
\label{fig:homeo3}
\end{figure}

For this, we let the homeostatic process run for 2000 time steps. In each of the time steps, the activity of every neuron is measured and the biases are increased such that the activity moves back to the baseline level with an adaptation rate of 0.01. After every time step, the performance of the network $Q$ is measured. The results of this process for the two types of connectivity are shown in Fig. \ref{fig:homeo3}. The performance at the beginning corresponds to the performance of the network when presented with a blank input. Then, in the early stages of the homeostatic process, a strong increase in the performance is visible. After around 1200 time steps, the performance saturates. 

We define the overall improvement $\Delta Q_{gain}$, visualized in figure \ref{fig:q_loss_q_gain}, as:
\begin{equation}
    \Delta Q_{gain} = Q_{hallucination} - Q_{blank}
\end{equation}

The final performance after 2000 steps of homeostasis $Q_{hallucination}$ is presented in the second last row of Table \ref{tab:results}. For both types of investigated connectivity, the homeostatic process leads to a significant increase in accuracy. In Fig. \ref{fig:homeo3}, the increase in performance from the initial state to the state after the homeostatic process $\Delta Q_{gain} \sim 0.2$ is shown.

\begin{figure}[h]
    \centering
    \includegraphics[width=0.45\textwidth]{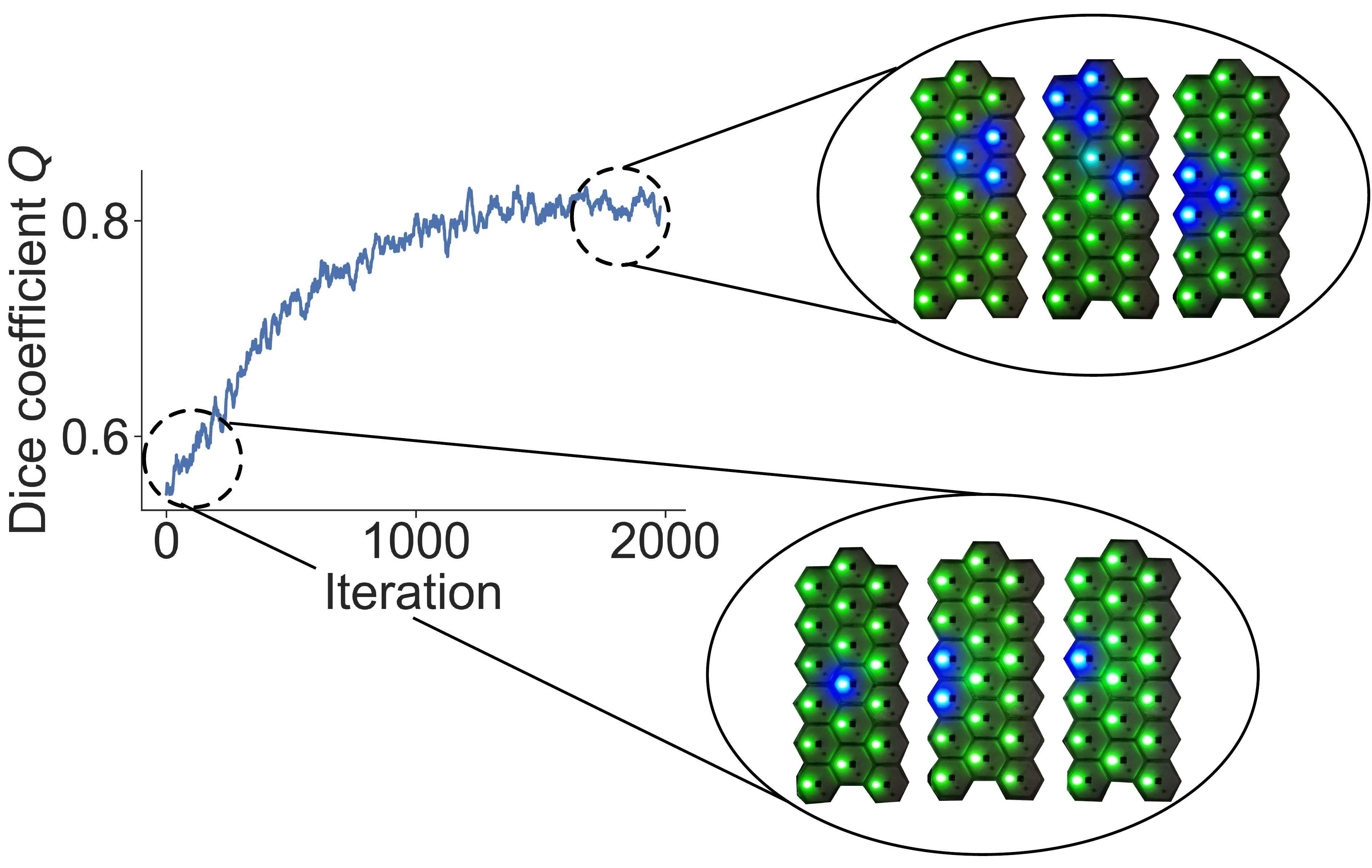}
    \caption{Demonstration of the reconstructed patterns given a sample trace during homeostasis for $\textit{ciruclar}$ connectivity. In the beginning, the reconstruction quality is rather poor and no patterns correspond to a training pattern. After applying homeostasis, several patterns correspond to the training samples.}
    \label{fig:homeo}
\end{figure}

Figure \ref{fig:homeo} then shows a sample trace for $\textit{circular}$ connectivity. Additionally, sample reconstructions are shown. While, before applying homeostatic mechanisms, the reconstructions are strongly perturbed, the decoded states at the final point clearly correspond to the trained patterns. Still, some of the decoded states correspond to corrupted versions of patterns.

\begin{figure}[h]
\centering   
\includegraphics[width=58mm]{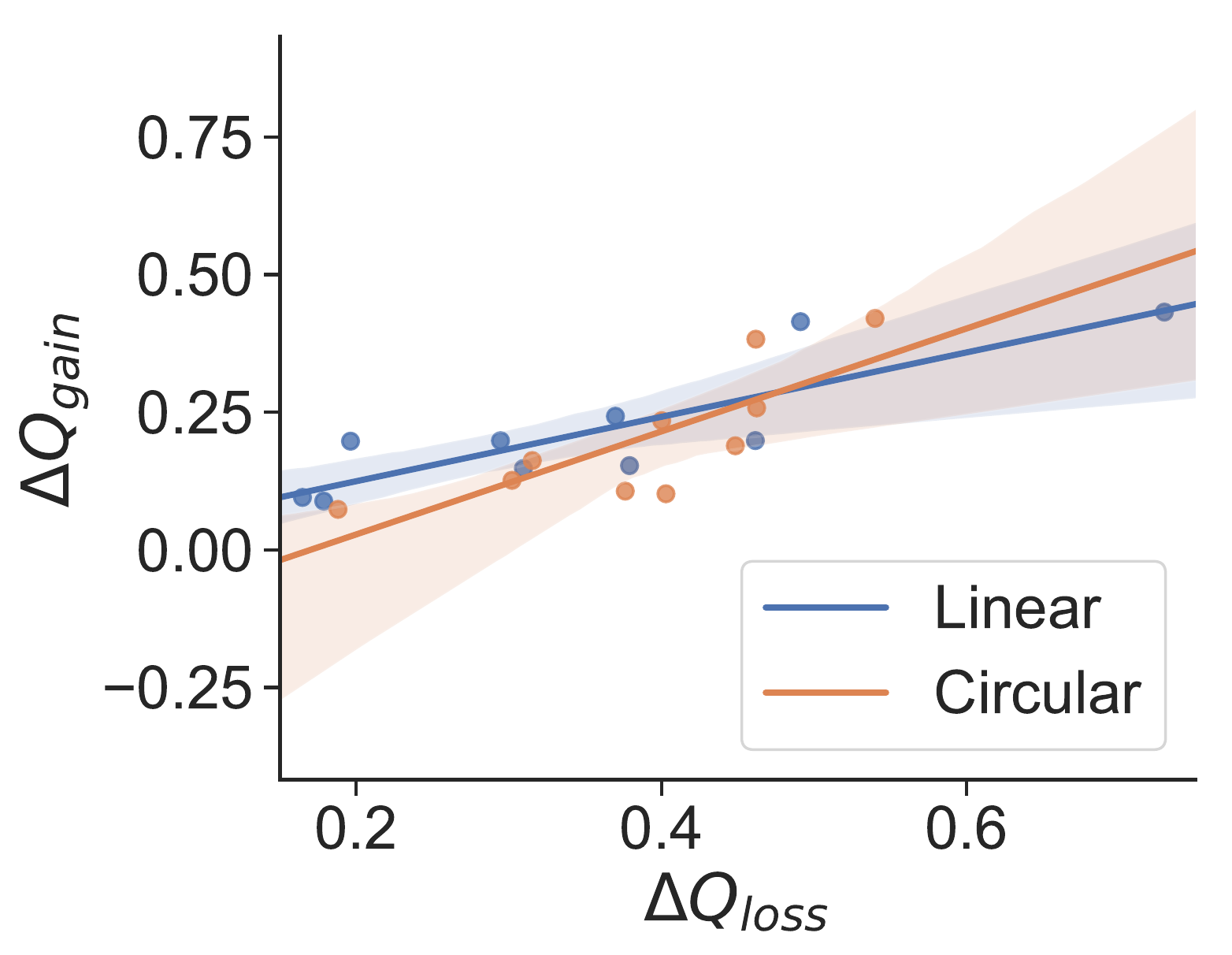}
\caption{Correlation between $Q_{loss}$ and $Q_{gain}$. On the $x$-axis, the difference in reconstruction performance between clamping to a training pattern or a blank pattern $Q_{loss}$ is shown. On the $y$-axis, the gain in performance through hallucinations $Q_{gain}$ is shown. The two traces correspond to the two types of connectivity namely $\textit{circular}$ (orange), and $\textit{linear}$ (blue). In all cases, there is a clear correlation (respective correlation coefficients: $\rho$=0.71, $\rho$=0.84). Each dot represents a separately trained DBM and is in itself an averaged value over 100 reconstruction samples of this particular DBM. The solid line is the least-squares regression line. The shaded area represents the standard deviation.}
\label{fig:corr}
\end{figure}

We further analyzed what affects the gain of performance through homeostasis. We found that there is a relation between the drop in performance $\Delta Q_{loss}$ and the gain through homeostasis $\Delta Q_{gain}$. This is shown in Fig. \ref{fig:corr} for both connectivity types. When the performance decreased strongly, the homeostatic mechanism will lead to a strong increase in performance afterwards (correlation coefficients, respectively, for the two lines: $\rho$=0.71, $\rho$=0.84). Thus, the difference in the quality of the reconstruction is particularly strong if the network suffered heavily from the blanked input pattern.

\section{Discussion}
\label{sec:discussion}
We suggest that tactile hallucinations can emerge as a consequence of homeostasis after an ill-formed input from the afferent pathway or other brain regions (corresponding to corrupted patterns) or through the complete lack of input (corresponding to blank patterns). Our results show that, when the network is clamped to one of the training patterns, the quality of the reconstruction is high, whereas it is low if the network is clamped to a corrupted or zeros pattern. Homeostatic mechanisms, modeled as an adaptation of the bias of the network, strongly increased the reconstruction quality. Thus, while homeostasis could be involved in the proper reconstruction or adaptation of sensory input when partial information is available, abnormal connectivity or dysfunctional homeostatic processes could induce tactile hallucinations.

\subsection{Medical applications}
Our results showed that homeostasis is a possible mechanism to induce hallucinations. In order to verify this approach, biological experiments would be required. A possible way to tackle this would be to match the timescales of homeostatic processes and the onset of hallucinations. In the phantom limb phenomenon, the onset has been reported to happen within the first 24 hours after amputation for half of the patients and within a week for another 25 \% of the patients \cite{weeks2010phantom}. This nicely fits the timescale of homeostatic mechanisms. Even though recent work has argued that there are fast homeostatic mechanisms on the timescale of seconds to minutes \cite{zenke2013synaptic}, most experimental work describes homeostasis to happen on a timescale of hours to weeks \cite{turrigiano2008self, turrigiano2017dialectic, watt2010homeostatic}.

If experimental work would further provide evidence for our hypothesis, it would thus be conceivable to make use of our approach in a medical application. In the cases where patients have unpleasant tactile hallucinations, alleviating the homeostatic mechanism could lead to a delayed onset of the hallucinations. Instead of letting the neurons recover previous baseline activity, medical interventions could keep the activity levels in the corresponding brain region low. This inference could happen in several ways, for example through drugs or targeted electrical stimulation.

\subsection{Technical applications}
The presented neuro-inspired framework can inspire technical systems beyond medical applications. DBMs have frequently been used as a method for feature extraction in classification tasks, where the features in the hidden layer are fed into a classifier \cite{decomparison}. In a robotics setting, homeostasis could then be used as an adaptive artificial intelligence mechanism. A general problem of artificial intelligence is domain adaptation, meaning that the network would perform badly in a different domain than the one it was trained on. In this work, we showed that keeping the learned weights constant and only changing the biases even in a non-selective way can recover the representations formed in the latent layers. Thus, even in non-optimal conditions, proper features could be selected by an adaptive DBM. As an example, the robot would be trained to perform a certain task under daylight conditions. However, when tested in a dim-light condition, we would expect it to perform worse. In that case, we could apply homeostasis to the DBM to recover the baseline activity that the classifier is familiar with. Furthermore, the generative nature of the proposed approach suits for flexible body perception and adaptation in robotics as an alternative method of predictive coding differential equations \cite{lanillos2018adaptive}.

\subsection{Technical challenges and limitations}
While our work provides a general framework for tactile hallucinations, it makes several, partly biologically implausible assumptions. Firstly, all our patterns were binary, while biological skin-receptors can have graded firing rates. A common approach is to either use tuning curves at the input layer to encode non-binary sensory information or to code the activity into a population of neurons representing a single skin cell \cite{salinas1994vector, franzen1991neural}. Secondly, while humans have millions of somatosensory receptors, we use only a small skin-patch. For further increasing the role of the spatial structure, a larger skin patch could be mounted on a robotic limb to allow for a natural interaction with the robot. As, in most cases, such interaction covers only a small fraction of all active skin cells, we used sparse patterns (see Fig. \ref{fig:train_data}). In comparison to dense patterns, these sparser patterns seemed to stabilize the performance of the network and of the homeostatic mechanisms. Besides, only by pre-training the DBM with a DBN, the system was able to learn the proper latent space representations. Conversely, training the RBM directly with $k$-step contrastive divergence was highly unstable and produced reasonable results only in a fraction of all trials.

Lastly, our approach has only used a small dataset. This does not resemble biology, where the input patterns are samples from a diverse distribution. This issue could generally be resolved by collecting more samples from a larger skin patch, as described in the paragraph above. In this case, the training data would then also come from actual interactions with a robot, allowing for diversity and realism in the dataset.

Further analysis should be performed on the spatial structure of the skin cells. In the two cases, using $\textit{circular}$ or $\textit{linear}$ receptive fields, the input matrix is linearized. In somatosensation, however, the input patterns have a highly ordered spatial structure. The effects of this spatial structure could be investigated by applying two-dimensional receptive fields or learning the spatial connectivity structure \cite{hoffmann2017robotic}. These receptive fields would lead to localized processing of the input data. Our work has already shown that constraining the connectivity has effects on the quality of the decoding and the strength of the hallucinations (Fig. \ref{fig:homeo3}). When moving towards more and more realistic tactile patterns, local processing and constrained connectivity could thus play a major role.

\section{Conclusion}
\label{sec:conclusion}
We presented a framework for modelling tactile hallucinations in an artificial skin. Overall, this work shows a possible role of homeostatic processes in the emergence of hallucinations. While homeostasis has been linked to many crucial parts of network function, we here showed that it could lead to the formation of meaningful latent representations without actual input. On the one hand, our model allows computational neuroscientists to investigate the potential role of different parameters such as the severity of the neurological disease (by modifying the corrupted input pattern). On the other hand, this framework also allows for more targeted experiments in order to elucidate the emergence of hallucinations.

\bibliographystyle{IEEEtran}
\bibliography{cbs2019}

\appendix

\subsection{Boltzmann Machines}
In general, all of the described neural network architectures are generative models. This means that, after a training phase, in which input patterns are being presented to the network, the network has learned the probability distribution of the underlying dataset. The network is thus able to sample from this probability distribution and produce samples that are consistent with the ones from the dataset.

In order to model hallucinations, this work leverages a Deep Boltzmann Machine (DBM) pre-trained with a Deep Belief Network (DBN). We will hence start by describing Restricted Boltzmann Machines (RBMs), the building blocks of DBMs and DBNs, and then go on to describing DBMs and DBNs themselves. After that, we will explain the common training algorithm for all these architectures, namely contrastive divergence. Lastly, we will describe how we sample from the DBM.

\subsection{Restricted Boltzmann Machines}
A Restricted Boltzmann Machine consists of two layers, termed the visible layer $v$ and the hidden layer $h$. The weights between the two layers are bidirectional and symmetric, meaning that they are shared between the forward and the backward pass. The neurons of all Boltzmann Machines are binary and stochastic. The probability for a neuron in the hidden layer $h_j$ (index $_j$ denotes the $j$-th neuron) to be one is given by the sigmoid of the sum over the weighted inputs from the visible layer:
\begin{equation}
    p(h_j^{(1)}|v) = \sigma\bigg(\sum_{i}v_i w_{ij} + b_i^{(1)}\bigg),
\end{equation}
where $v$ denotes the visible layer activations, $w_{ij}$ is the weight from neuron $j$ to $i$, and $b_i$ is the bias term of the $i$-th neuron. Moreover, $\sigma(\cdot)$ denotes the sigmoid function $1/(1+\exp(-x))$. By minimizing the free energy
\begin{equation}
    E(v, h, \theta) = -b^T v - c^{T}h - v^T W h, 
\end{equation}
of the states $v, h$ with respect to all parameters $\theta$, which contains the weights $W$ and the biases $b, c$, the network learns to represent the features of the visible layer in the hidden layer and thereby allows for sampling from the same distribution as the input data \cite{smolensky1986information, hinton2002training}.

\subsection{Deep Boltzmann Machines}
\label{sec:DBM}
Deep Boltzmann Machines (DBMs) are Markov random fields. Unlike Deep Belief networks, the intermediate layers of a DBM receive input from both its subsequent layers. Hence, the activation of a neuron in the first hidden layer can be computed as:
\begin{equation}
    p(h_j^{(1)}|h_j^{(2)}, v) = \sigma\bigg(\sum_{i}v_i w_{ij}^{(1)} + \sum_{k}h_k^{(2)} w_{kj}^{(2)} + b_i^{(1)}\bigg ),
\end{equation}
where the subscript $^1$ denotes the first layer. Like RBMs, DBMs learn by minimizing the free energy. In a DBM with two hidden layers, the free Energy $E$ is defined as
\begin{equation}
  \begin{multlined}
    E(v, h^{(1)}, h^{(2)}, \theta) = -b^T v - c^{(1)T}h^{(1)} -\\ c^{(2)T}h^{(2)} - v^T W^{(1)} h^{(1)} - h^{(1)T} W^{(2)} h^{(2)}.
    \end{multlined}
\end{equation}
As described above, DBMs also have stochastic activations. The probability of a certain pattern to be observed in the DBM can be calculated from its free energy:
\begin{equation}
    p(v, \theta) = \frac{1}{Z} \exp(-F(v)),
    \label{eq:prob}
\end{equation}
where $Z$ is the partition function defined as
\begin{equation}
    Z = \sum_{x} \exp(-F(x)),
\end{equation}
with $x$ denoting all possible states. The partition function hence acts as a normalization factor. Moreover, the function $F(v)$ denotes
\begin{equation}
    F(x) = - \log\big(\sum_h \exp(-E(x))\big).
    \label{eq:F}
\end{equation}

\subsection{Deep Belief Networks}
\label{sec:DBN}
Unlike DBMs, Deep Belief Networks (DBNs) represent several stacked RBMs, where each layer learns the features of the more shallow layer. DBNs have directed connections which implies that the training happens in a greedy, layer-wise way. In practice, we feed the data into the visible layer and use persistent contrastive divergence, as described in section \ref{sec:trBM} to train the first hidden layer. Then, the second hidden layer is trained by using the states of the first hidden layer that had been computed via the visible layer.

\subsection{Training Boltzmann Machines}
\label{sec:trBM}
Training in Boltzmann Machines happens by minimizing the difference in free energies of the data and the model, which can be described in two steps. The algorithm that implements this is called persistent contrastive divergence. The Boltzmann Machine is first clamped to the data and the hidden layers are being updated. Through this, the network is forced into a state that does generally not correspond to an energy minimum yet, but rather a state that represents the input pattern. Once the network has reached its equilibrium, we calculate the free energy of the state of the network. It has been shown that a single update loop is sufficient to successfully train Boltzmann Machines \cite{hinton2002training}. A gradient descent approach will then minimize this free energy and thereby maximize the probability of the data (the clamped tactile pattern). After this, the network is run freely, without any clamped layers (again, a single update loop is sufficient \cite{hinton2002training}). In this process, the states of the network end up in a local energy minimum. We can use gradient ascent to maximize the free energy of the model for this state in order to decrease the probability of unwanted patterns. Note that, in practice, these two update states are performed simultaneously. We described them as separate processes for clarity. The resulting learning rule can be written as
\begin{equation}
    \Delta w^{0} = \lambda \big((x^{(0)} x^{(1)T})_{data} - (x^{(0)} x^{(1)T})_{model}\big)
\end{equation}
with learning rate $\lambda$. Note that this can be identified as a difference between two Hebbian learning equations, and hence a local learning rule. Hebbian and anti-Hebbian learning have been found in many regions of the brain \cite{caporale2008spike, foldiak1990forming}.

\subsection{Sampling from the Boltzmann Machines}
Sampling from Boltzmann Machines happens by feeding random inputs into the network. We then update the states of all layers. Through this, the state of the network moves into a local minimum of the free energy. Which local minimum the network moves into is defined by the starting condition, i.e. the random input pattern. For a well trained Boltzmann machine, the state corresponding to a free-energy-minimum represents a sample from the data distribution. Thus, we can take the values of the visible layer and use them as our sample.

\end{document}